\def\year{2018}\relax
\documentclass[letterpaper]{article} 
\usepackage{aaai18}  
\usepackage{times}  
\usepackage{helvet}  
\usepackage{courier}  
\usepackage{url}  
\usepackage{graphicx}  
\usepackage{tabularx}
\usepackage{color}
\usepackage{makecell}
\newcolumntype{?}{!{\vrule width 2pt}}
 
\begin{document}
%
\title{Co-Creative Level Design via Machine Learning}
\author{Matthew Guzdial, Nicholas Liao, and Mark Riedl\\
College of Computing\\
Georgia Institute of Technology\\
Atlanta, GA 30332\\
mguzdial3@gatech.edu, nliao7@gatech.edu, riedl@cc.gatech.edu\\
}
\maketitle
\begin{abstract}
Procedural Level Generation via Machine Learning (PLGML), the study of generating game levels with machine learning, has received a large amount of recent academic attention.
For certain measures these approaches have shown success at replicating the quality of existing game levels. However, it is unclear the extent to which they might benefit human designers.
In this paper we present a framework for co-creative level design with a PLGML agent.
In support of this framework we present results from a user study and results from a comparative study of PLGML approaches. 

\end{abstract}

\section{Introduction}

Procedural content generation via Machine Learning (PCGML) has drawn increasing academic interest in recent years \cite{summerville2017procedural}. 
In PCGML a machine learning model trains on some existing corpus of game content to learn a distribution over possible game content. 
New content can then be sampled from this distribution.
This approach has shown some success at replicating existing game content, particularly of game levels, according to user studies \cite{guzdial2016game} and quantitative metrics \cite{snodgrass2017learning,summerville2018learning}.
The practical application of PCGML approaches has not yet been investigated. 
One might naively suggest that PCGML could serve as a cost-cutting measure given its ability to generate new content that matches existing content. 
However, this requires a large corpus of existing game content. If designers for a new game produced such a corpus, they might as well use that corpus for the final game. 
Beyond this issue, a learned distribution is not guaranteed to contain a designer's desired output.

A co-creative framework could act as an alternative to asking designers to find desired output from a learned distribution . In a co-creative framework, also called mixed initiative, a human and AI partner work together to produce final content. In this way, it does not matter if an AI partner is incapable of creating some desired output alone. 

In this paper we propose an approach to co-creative PCGML for level design or Procedural Level Generation via Machine Learning (PLGML). In particular, we intend to demonstrate the following points: (1) existing methods are insufficient for co-creative level design, and (2) co-creative PLGML requires training on examples of co-creative PLGML or an approximation. In support of this argument we present results from a user study in which users interacted with existing PLGML approaches adapted to co-creation and quantitative experiments comparing these existing approaches to approaches designed for co-creation.

\section{Related Work}

The concept of co-creative PCGML has been previously discussed in the literature \cite{summerville2017procedural,zhuexplainable}, but no prior approaches or systems exist. Comparatively there exist many prior approaches to co-creative or mixed-initiative level design agents without machine learning \cite{smith2010tanagra,yannakakis2014mixed,deterding2017mixed}. Instead these approaches rely upon search or grammar-based approaches \cite{liapis2013sentient,shaker2013ropossum,baldwin2017mixed}. Thus these approaches require significant developer effort to adapt to a novel game.

\begin{figure*}[tb]
\centering
	\includegraphics[width=5in]{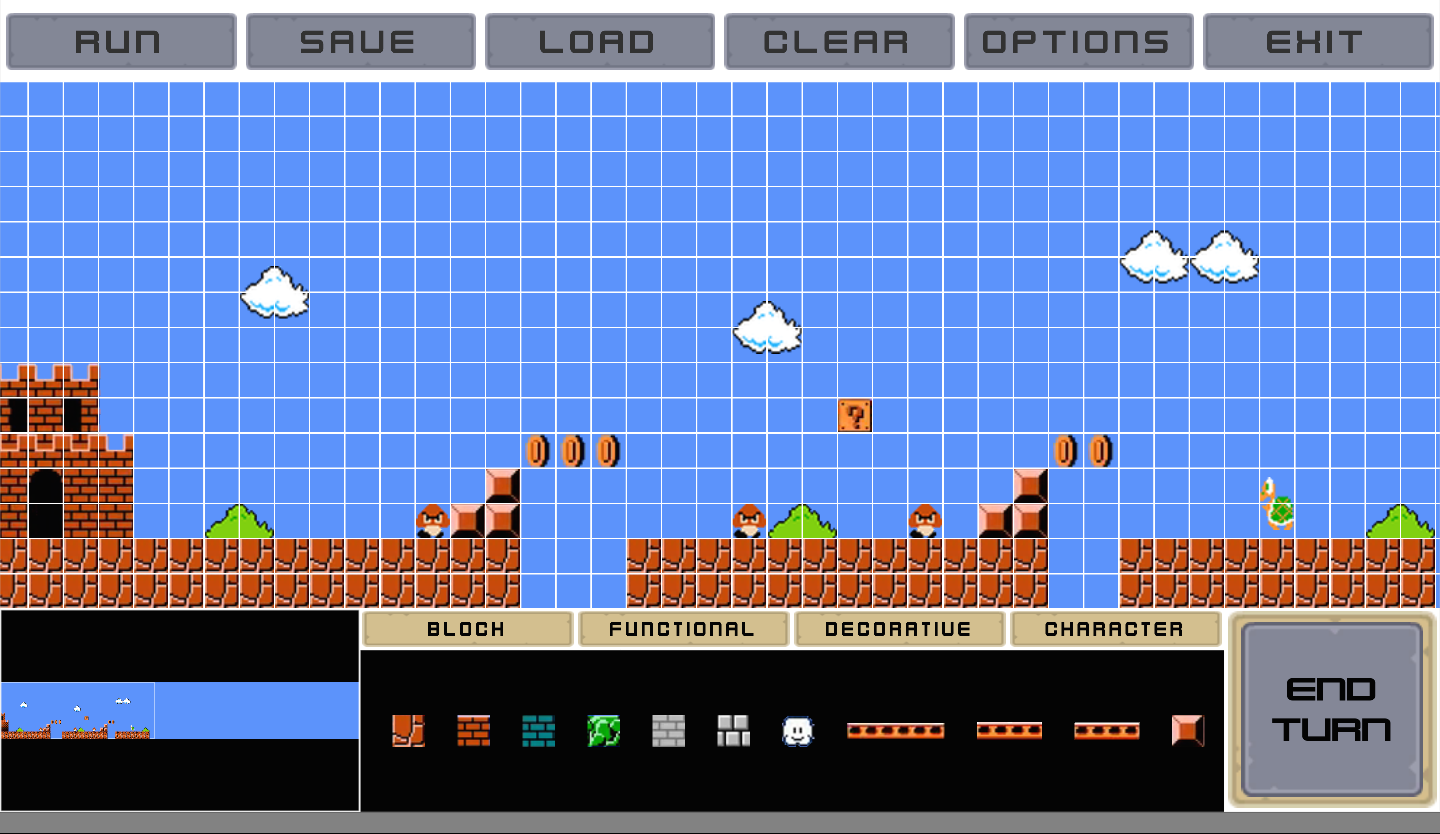}
	\caption{Screenshot of the Level Editor, reproduced from \cite{guzdial2017general}.}
	\label{fig:screenshot}
\end{figure*}

\begin{figure}[tb]
\centering
	\includegraphics[width=\columnwidth]{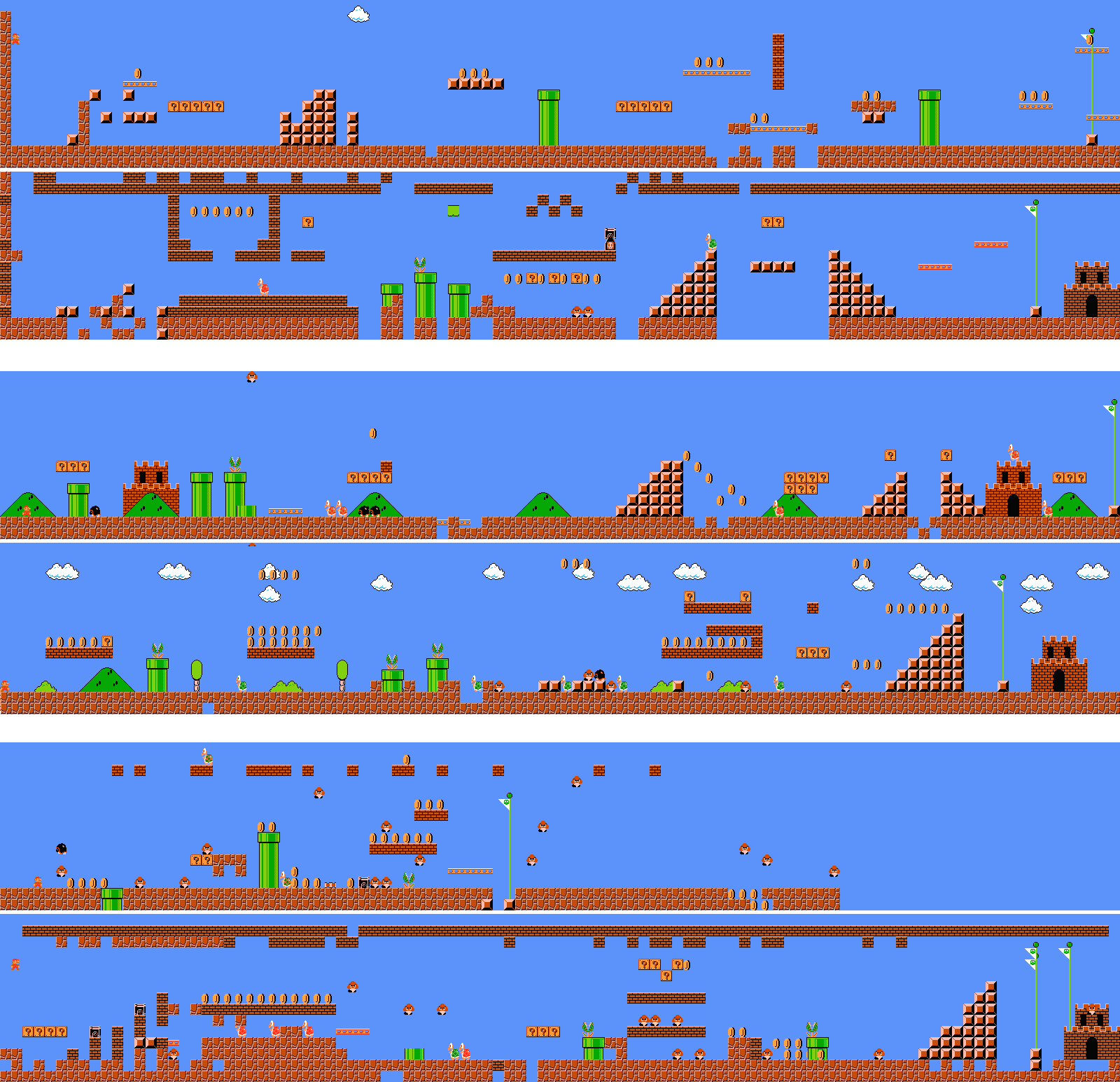}
	\caption{Examples of six final levels from our study, each pair of levels from a specific co-creative agent: Markov Chain (top), Bayes Net (middle), and LSTM (bottom). These levels were selected at random from the set of final levels, split by co-creative agent.}
	\label{fig:exampleLevels}
\end{figure}

\section{User Study}

As an initial exploration into co-creative level design via machine learning we conducted a user study. We began by taking existing procedural level generation via machine learning (PLGML) approaches and adapting them to co-creation. We call these adapted approaches AI level design partners. Our intention with these partners is to determine the strengths and weaknesses of these existing approaches when applied to co-creation and the extent to which these existing approaches are sufficient for this task. 

We make use of Super Mario Bros. as the domain for this study and later experiments given that all three of the existing PLGML approaches had previously been applied to this domain. Further, we anticipated its popularity would lead to more familiarity from our study participants. 

\subsection{Level Design Editor}

To run our user study we needed some Level Design Editor to serve as an interface between participants and the AI level design partners. For this purpose we made use of the editor from \cite{guzdial2017general}, which is publicly available online.\footnote{https://github.com/mguzdial3/Morai-Maker-Engine} We reproduce a screenshot of the interface from the paper in Figure \ref{fig:screenshot}. The major parts of the interface are as follows:
\begin{itemize}
\item The current level map in the center of the interface, which allows for scrolling side-to-side
\item A minimap on the bottom left of the interface, users can click on this to jump to a particular place in the level
\item A palette of level components or sprites in the middle of the bottom row
\item An ``End Turn'' button on the bottom right. By pressing this End Turn button the current AI level design partner is queried for an addition. A pop-up appears while the partner processes, and then its additions are added sprite-by-sprite to the main screen. The camera scrolls to follow each addition, so that the user is aware of any changes to the level. The user then regains control and level building continues in this turn-wise fashion.
\end{itemize}

\noindent
At any time during the interaction users can hit the top left ``Run'' button to play through the current version of the level. A backend logging system tracks all events, including additions and deletions and which entity (human or AI) was responsible for them.

\subsection{AI Level Design Partners}

For this user study we created three AI agents to serve as level design partners. Each is based on a previously published PLGML approach, adapted to work in an iterative manner to fit the requirements of our level editor interface. We lack the space to fully describe each system but cover a high-level summary of the approaches and our alterations below.

\begin{itemize}
\item \textbf{Markov Chain: } This approach is a Markov chain based on Snodgrass and Ontan{\'o}n \shortcite{snodgrass2014experiments}, based on Java code supplied by the authors. It trains on existing game levels by deriving all 2-by-2 squares of tiles and deriving probabilities of a final tile from the remaining three tiles in the square. We made use of the same representation as that paper, which represented elements like enemies and solid tiles as equivalent. To convert this representation to the editor representation we applied rules to determine the appropriate sprite from the solid tile class based on its position and chose randomly from available enemies for the enemy class (with the stipulation that flying enemies could only appear in the air). Otherwise, our only variation from this baseline was to limit the number of new generated tiles to a maximum of thirty per turn.
\item \textbf{Bayes Net: } This approach is a probabilistic graphical model or hierarchical Bayesian network based on Guzdial and Riedl \shortcite{guzdial2016game}. It derives shapes of sprite types and samples from a probability of relative positions to determine the next sprite shape to add and where. This approach was originally trained on gameplay video, thus we split each level into a set of frame-sized chunks, and generated an additional shape for each chunk. This approach was already iterative and so naturally fit into the turn-based level design format. We do not limit the number of additions, but the agent only made additions when there was a sufficient probability, and thus almost always produced fewer additions than the other agents.
\item \textbf{LSTM: } This approach is a Long Short Term Memory Recurrent Neural Network (LSTMRNN or just LSTM) based on Summerville and Mateas \shortcite{summerville2016super}, recreated in Tensorflow from the information given in the paper and training data supplied by the authors. It takes as input a game level represented as a sequence and outputs the next tile type. We modified this approach to a bidirectional LSTM given it was collaborating and not just building a level from start to end. We further modified the approach to only make additions to a 65-tile wide chunk of the level, centered on the user's current camera placement in the editor. As with the Markov Chain we limited the additions to 30 at most, and converted from the agent's abstract representation to the editor representation according to the same process.
\end{itemize}

\noindent
We chose these three approaches as they represent the most successful prior PLGML approaches in terms of depth and breadth of evaluations. Further, each approach is distinct from the other two. For example, each approach has a difference in terms of local vs. global reasoning, with the Markov Chain being hyper-local (only generating based on a 2x2 square) to the much more global LSTM approach which reads in almost the entirety of the current level. Notably, because all three approaches were previously used for autonomous generation, the agents could only make additions to the level, never any deletions. We did not put any effort to including deletions in order to minimize the damage the agent could cause to a user's intended design of a level.

\subsection{Study Method}

Each study participant went through the same process. First, they were given a short tutorial on the level editor and its function. They then interacted with two distinct AI partners back-to-back. The partners were assigned at random from the three possible options. During each interaction, the user was assigned one of two possible tasks, either to create an above ground or below ground level. We supplied two optional examples of the first two levels of each type taken from the original Super Mario Bros.. This leads to a total of twelve possible conditions in terms of pair of partners, order of the pair, and order of the level design assignments.

Participants were given a maximum of fifteen minutes for each task, though most participants finished well before then. Participants were asked to press the ``End Turn'' button to interact with their AI partner at least once. Those who did not do so had their results thrown out. 

After both rounds of interaction participants took a brief survey in which they ranked the two partners they interacted with in terms of fun, frustration, challenge to work with, the partner that most aided the design, the partner that lead to the most surprising or valuable ideas, and which of the two partners the participant would most like to use again. We also gave participants the option to leave a comment reflecting on each agent. The survey ended by collecting demographic data including experience with level design, Super Mario Bros., games in general, the participant's gender (we collected gender in a free response field), and age.  

\subsection{Results}

\begin{table*}[tb]
\begin{center}
\caption{A table comparing the ratio by which each system was ranked first or second.}
\begin{tabular}{ |l|c|c|c|c|c|c|c| } 
 \hline
 & Most Fun & Most Frustrating & Most Challenging & Most Aided & Most Creative & Reuse \\ 
 \hline
 Markov Chain & 33:23 & 26:30 & 29:27 & 30:26 & 33:23 & 32:24\\
 \hline
 Bayes Net & 27:29 & 26:30 & 20:36 & 31:25 & 29:27 & 28:28 \\
 \hline
LSTM & 24:32 & 32:24 & 35:21 & 23:33 & 22:34 & 24:32 \\ 
 \hline
\end{tabular}
\end{center}
\label{tab:subjectResults}
\end{table*}

In this subsection we discuss an initial analysis of the results of our user study. Overall 91 participants took part in this study. However, of these seven participants did not interact with one or both of their partners, and we removed them from our final data. The remaining 84 participants were split evenly between the twelve possible conditions, meaning a total of seven participants for each condition. 

62\% of our respondents had previously designed Mario levels at least once before. This is likely due to prior experience playing Mario Maker, a level design game/tool released by Nintendo on the Wii U. Our subjects were nearly evenly split between those who had never designed a level before 26\%, designed a level once before 36\%, or had designed multiple levels in the past 38\%. All but 7 of the subjects had previously played Super Mario Bros., and all the subjects played games in general regularly. 

Our first goal in analyzing our results was to determine if the level design task (above or underground) mattered and if the ordering of the pair of partners mattered. We ran a one-way repeated measures ANOVA and found that neither variable lead to any significance. Thus, we can safely treat our data as having only three conditions, dependent on the pair of partners each subject interacted with. 

We give the ratio of first place to second place rankings for each partner in Table 1. Therefore one can read the results as the Markov Chain agent being generally preferred, though more challenging to use. Comparatively, the Bayes net agent was considered less challenging to use, but also less fun, with subjects less likely to want to reuse the agent. The LSTM on the other hand had the worst reaction overall. 

The ratio of ranking results would seem to indicate a clear ordering of the agents. However, this is misleading. We applied the Kruskal Wallis test to the results of each question and found it unable to reject the null hypothesis that all of the results from all separate agents arose from the same distribution. This indicates that in fact the agents are too close in performance to state a significant ordering. In fact, many subjects greatly preferred the LSTM agent over the other two, stating that it was ``Pretty smart overall, added elements that collaborate well with my ideas'' and ``This agent seemed to build towards an `idea' so to speak, by adding blocks in interesting ways''. 

\subsection{User Study Results Discussion}

These initial results of our user study do not indicate a clearly superior agent. Instead, they suggest that individual participants varied in terms of their preferences. This matches our own experience with the agents. When attempting to build a very standard Super Mario Bros. level, the LSTM agent performed well. However, as is common with deep learning methods it was brittle, defaulting to the most common behavior (e.g. adding grounds or blocks) when confronted with unfamiliar input. In comparison the Bayes net agent was more flexible, and the Markov Chain agent more flexible still, given its hyper-local reasoning.

We include two randomly selected levels for each agent in Figure \ref{fig:exampleLevels}. They clearly demonstrate some departures from typical Super Mario Bros. levels, meaning none of these levels could have been generated by any of these agents. Given this, and the results of the prior section, we have presented some evidence towards the first part of our argument, that existing methods are insufficient to handle the task of co-creative level design. By which we mean, no existing agents are able to handle the variety of human level design or human preferences when it comes to AI agent partners. We will present further evidence towards this and the second point in the following sections.

\section{Proposed Co-Creative Approach}

The results of the prior section indicate a need for an approach designed for co-creative PLGML instead of adapted from autonomous PLGML. In particular, given that none of our existing agents were able to sufficiently handle the variety of participants, we expect instead a need for an ideal partner to either more effectively generalize across all potential human designers or to adapt to a human designer actively during the design task. We present a proposed architecture based on the results of the user study, and present both pre-trained and active learning variations to investigate these possibilities. 

\subsection{Dataset}

For the remainder of this paper we make use of the results of the user study as a dataset. In particular, as stated in the Level Design Editor subsection, we logged all actions by both human and AI agent partners. These logs can be considered representations of the actions taken during each partner's turns. We also have final scores in terms of the user ranking. These final scores could serve as reward or feedback to a supervised learning system, however, we would ideally like some way to assign partial credit to all of the actions the AI agent took to receive those final scores. Towards this purpose we decided to model this problem as a general, semi-Markov Decision Process (SMDP) with concurrent actions as in \cite{rohanimanesh2003learning}. 

Our SMDP with concurrent actions is from the AI partner's perspective, given that we wish to use it to train a new AI partner. It has the following components:
\begin{itemize}
\item \textbf{State: } We represent the level at the end of each human user turn as the state.
\item \textbf{Action: } Each single addition by the agent per turn then becomes a primary action, with the total turn representing the concurrent action.
\item \textbf{Reward: } For the reward we make use of the Reuse ranking, as it represents our desire that the agent be helpful and usable first and foremost. In addition, we include a small negative reward (-0.1) if the user deletes an addition made by the AI partner. We make use of a $\gamma$ value of 0.1 in order to determine partial credit across the sequences of AI partner actions.
\end{itemize}

\noindent
Due to some network drops, some of the logs for our study were corrupted. Thus we ended up with 122 final sequences from our logs. We split this dataset into a 80-20 train-test split by participant, ensuring that our test split only included participants with both logs from both interactions uncorrupted. Thus we had the logs of 11 participants held out for testing purposes.

We further divided each state-action-reward triplet such that we represent each state as a 40x15x32 matrix and each action as a 40x15x32 matrix. The state represents a screens worth of the current level (40x15), and the action represents the additions made over that chunk of level. The 32 in this case is a one-hot encoding of sprites, based on the 32 possible sprites in the editor's sprite palette. We did this in order to further increase the amount of training data. This lead to a total of 1501 training samples and 242 test samples.

\subsection{Architecture}

From our user study we found that local coherency (Markov Chain) tended to outperform global coherency (LSTM). Thus for a proposed co-creative architecture we chose to make use a Convolutional Neural Network (CNN). A CNN is capable of learning local features that impact decision making, and to replicate those local features for generation purposes. Further, they have shown success in approximating the Q-table in more traditional deep reinforcement learning applied to game playing \cite{mnih2013playing}.

We made use of a three layer CNN, with the first layer having 8 4x4 filters, the second layer having 16 3x3 filters, and the final layer having 32 3x3 filters. The final layer is a fully connected layer followed by a reshape to place the output in the form of the action matrix (40x15x32). Each layer made use of leaky relu activation, meaning that each index of the final matrix could vary from -1 to 1. We made use of mean square loss and adam as our optimizer, with the network built in Tensorflow \cite{abadi2016tensorflow}. We trained this model to the point of convergence in terms of training set error.

\subsection{Pretrained Evaluation}

\begin{table}[tb]
\begin{center}
\caption{A table comparing the summed reward each agent receives on the test data.}
\begin{tabular}{ |l?c|c|c|c|c| } 
 \hline
 participant & Ours & SMB & MC & GR & LSTM \\ 
 \hline
 0 & 1.45 & 7.34 & \textbf{10.0} & 0.00 & 10.0 \\
 \hline
  1 & \textbf{1.32} & -4.63 & -4.00 & -1.00 &  -6.00\\
 \hline
2 & 0.00 & 0.00 & 0.00 & 0.00 & 0.00 \\
 \hline
3 & \textbf{-0.53} & -1.57 & 0.00 & 0.00 & -3.00 \\
 \hline
4 & 0.01 & \textbf{0.31} & 0.00 & 0.00 & 0.00 \\
 \hline
5 & \textbf{5.50} & 1.36 & 0.00 & 0.00 & 1.00 \\
 \hline
6 & \textbf{0.29} & -0.07 & 0.00 & 0.00 & 0.00 \\
 \hline
7 & 0.10 & 1.00 & \textbf{2.00} & 0.00 & 1.00 \\
 \hline
8 & \textbf{-0.14} & -10.1 & -60.1 & 0.00 & -40.2 \\
 \hline
9 & 3.85 & \textbf{14.0} & 0.00 & 0.00 & -1.10 \\
 \hline
10 & \textbf{-3.01} & -5.89 & 0.00 & 0.00 & 0.00 \\
 \hline
 \hline
Avg \% & \textbf{53.9} & 0.8 & -0.6 & -0.0 & -0.5 \\
 \hline
\end{tabular}
\end{center}
\label{tab:pretrainedResults}
\end{table}

For our first evaluation we compared the total reward accrued on the test set across our 242 withheld test samples. In comparison we make use of four baselines, the three existing agents and one variation on our approach. 

For the variation on our approach, we instead trained on a dataset created from the existing levels of Super Mario Bros. (SMB), represented in our SMDP format. To accomplish this, we derived all 40x15x32 chunks of SMB levels. We then removed all sprites of each single type from that chunk, which became our state, with the action being the addition of those sprites. We made the assumption that each action should receive a reward of 1, given that it would lead to a complete Super Mario Bros. level. 

This evaluation can be understood as running these five agents (our approach, the SMB variation, and the three already introduced agents) through a simulated interaction with the held out test set of eleven participants. This is not a perfect simulation, given that we cannot estimate reward without user feedback. However, given the nature of our reward function, actions that we cannot assign reward to will receive 0.0. This makes the final amount of reward each agent receives a reasonable estimate of how each person might respond to the agent.

The second claim we made was that co-creative PLGML requires training on examples of co-creative PLGML or an approximation. Thus our proposed approach can be considered the former of these two and the variation of our approach trained on the Super Mario Bros. dataset the latter. If these two approaches outperform the three baselines we will have evidence for this, and our first claim that existing PLGML methods were insufficient for co-creation.

\subsection{Pretrained Evaluation Results}

We summarize the results of this evaluation in Table 2. The columns represent in order the results of our approach, the SMB-trained variation of our approach, the Markov Chain baseline, the Bayes net baseline, and the LSTM baseline. The rows represent the results separated by each participant in our test set. We separate the results in this way given the variance each participant displayed, and since the total possible reward would depend upon the number of interactions, which differed between participants. Further, each participant must have given both a positive and negative final reward (ranking agents first and second in terms of reuse). Due to this reason we present the results in terms of summed reward per-participant. Thus, higher is better. It is possible for an agent to achieve a negative reward if it places items that the participant removed or that correspond with a final -1 reward. Further, it is possible to end up with a summed reward of 0 if the agent takes actions that we cannot assign any reward. For example, if we know that a human participant doesn't want an enemy, but the agent adds a pipe. We cannot estimate reward in this case. Finally, it is possible to end with a summed reward much larger than 1.0 given a large number of actions that encompassed a large amount of the level (thus many 40x15x32 testing chunks). The final row indicates the average percentile performance our of the maximum possible reward for each participant, since once normalized we can average these results to present them in aggregate. 

The numbers in Table 2 cannot be compared between rows given how different the possible rewards and actions of each participant was. However, we can compare between columns. For the final row, our approach and the SMB variation are the only two approaches on average to receive positive reward. We note that the Markov Chain partner does well for some individuals, but overall has a worse performance than the LSTM agent. The Bayes net agent may appear to do better, but this is largely because it either predicted nothing for each action or something for which the dataset did not have a reward. We note that participant 2 in the Table received a summed reward of 0.0 for all the approaches, but this is because that participant only interacted with their two agents once and did not make any deletions.

\subsection{Active Evaluation}

The prior evaluation demonstrates that by training on a dataset or approximated dataset of co-creative interactions one can outperform machine learning approaches trained to autonomously produce levels. This suggests these approaches do a reasonable job of generalizing across the variety of interactions in our training dataset. However, if designers vary extremely from one another, generalizing too much between designers will actively harm a co-creative agent's potential performance. This second comparative evaluation tests if this is the case.

For this evaluation we create two active learning variations of our approach. For both, after making a prediction and receiving reward for each test sample we then train on that sample for one epoch. In the first, we reset the weights of our network to the final weights after training on our training set after every participant (we call this variation ``Episodic"). In the second, we never reset the weights, allowing the agent to learn and generalize more from each participant it interacts with (We call this variation ``Continuous"). If it is the case that user designs vary too extremely for an approach to generalize between them, then we would anticipate ``Continuous" to do worse, especially as it gets to the end of the sequence of participants.

\begin{table}[tb]
\begin{center}
\caption{A table comparing two variations on an active learning version of our agent.}
\begin{tabular}{ |l?c|c|c|} 
 \hline
 participant & Ours & Episodic & Continuous \\ 
 \hline
 0 & 1.45 & \textbf{1.47} & \textbf{1.47}  \\
 \hline
  1 & \textbf{1.32} & -11.7 & -10.1 \\
 \hline
2 & 0.00 & 0.00 & 0.00 \\
 \hline
3 & -0.53 & 0.94 & \textbf{1.08} \\
 \hline
4 & \textbf{0.01} & -0.05 & -0.25\\
 \hline
5 & \textbf{5.50} & \textbf{5.50} & -7.55 \\
 \hline
6 & \textbf{0.29} & \textbf{0.29} & 0.04  \\
 \hline
7 & \textbf{0.10} & \textbf{0.10} & -0.04 \\
 \hline
8 & -0.14 & \textbf{5.22} & 0.42  \\
 \hline
9 & 3.85 & \textbf{42.7} & 41.0 \\
 \hline
10 & \textbf{-3.01} & -3.76 & -4.62 \\
 \hline
 \hline
Avg \% & 53.9 &  \textbf{56.6} & 53.1 \\
 \hline
\end{tabular}
\end{center}
\label{tab:activeResults}
\end{table}

\subsection{Active Evaluation Results}

We summarize the results of this evaluation in Table 3. We replicate the results of the non-active learning version of our approach from Table 2. Overall, these results support out hypothesis. The average percentile of the maximum possible reward increased by roughly three percent from the non-active version to the episodic active learner, and decreased by roughly a percentage point for the continuous active learner. The continuous active learner did worse than either the episodic active learner or our non-active learner for six of the eleven participants. This indicates that participants do tend to vary too much to generalize between, at least for our current representation.

Overall, it appears that some participants were more or less easy to learn from. For example, participants 1, 4, and 10 all did worse with agents attempting to adapt to them during the simulated interaction. However, participants 8 and 9 both seemed well-suited to adaption given that their scores increased over ten times from the non-active learner. This follows from the fact that these two participants had the second most and most interactions respectively across the test participants. This suggests the ability for these agents to adjust to a human designer given sufficient interaction.

\section{Discussion and Limitations}

In this paper we presented results towards an argument for co-creative level design via machine learning. We presented evidence from a user study and two comparative experiments that (1) current approaches to procedural level generation via machine learning are insufficient for co-creative level design and (2) that co-creative level design requires training on a dataset or an approximated dataset of co-creative level design. In support, we demonstrate that no current approach significantly outperforms any of the remaining approaches, and in fact that users are too varied for any one model to meet an arbitrary user's needs. Instead, we anticipate the need to apply active learning to adapt a general model to particular individuals.

We present a variety of evidence towards our stated claims. However, we note that we only present evidence in the domain of Super Mario Bros.. Further, while our comparative evaluations had strong results, these can only be considered simulations of user interaction. In particular, our simulated test interactions essentially assume users will create the same final level, no matter what the AI partner does. To fully validate these results we will need to run a new user study. We anticipate running a follow up study in order to verify these results.

Beyond a follow-up user study, we also hope to investigate ways of speeding up the process of creating co-creative level design partners. Under the process described in this paper, one would have to run a 60+ user study with three different naive AI partners every time you wanted a co-creative level design partner for a new game. We plan to investigate transfer learning and other ways to approximate co-creative datasets from existing corpora. Further, we anticipate a need for explainable AI in co-creative level design to help the human partner give appropriate feedback to the AI partner. 

\section{Conclusions}

We introduce the problem of co-creative level design via machine learning. This represents a new domain of research for Procedural Level Generation via Machine Learning (PLGML). In a user study and two comparative evaluations we demonstrate evidence towards the claim that existing PLGML methods are insufficient to address co-creation, and that co-creative AI level designers must train on datasets or approximated datasets of co-creative level design.

\section{Acknowledgements}

This material is based upon work supported by the National Science Foundation under Grant No. IIS-1525967. This work was also supported in part by a 2018 Unity Graduate Fellowship.

\bibliographystyle{aaai}
\bibliography{aaai}

\begin{thebibliography}{}

\bibitem[\protect\citeauthoryear{Abadi \bgroup et al\mbox.\egroup
  }{2016}]{abadi2016tensorflow}
Abadi, M.; Barham, P.; Chen, J.; Chen, Z.; Davis, A.; Dean, J.; Devin, M.;
  Ghemawat, S.; Irving, G.; Isard, M.; et~al.
\newblock 2016.
\newblock Tensorflow: A system for large-scale machine learning.
\newblock In {\em OSDI}, volume~16,  265--283.

\bibitem[\protect\citeauthoryear{Baldwin \bgroup et al\mbox.\egroup
  }{2017}]{baldwin2017mixed}
Baldwin, A.; Dahlskog, S.; Font, J.~M.; and Holmberg, J.
\newblock 2017.
\newblock Mixed-initiative procedural generation of dungeons using game design
  patterns.
\newblock In {\em Computational Intelligence and Games (CIG), 2017 IEEE
  Conference on},  25--32.
\newblock IEEE.

\bibitem[\protect\citeauthoryear{Deterding \bgroup et al\mbox.\egroup
  }{2017}]{deterding2017mixed}
Deterding, C.~S.; Hook, J.~D.; Fiebrink, R.; Gow, J.; Akten, M.; Smith, G.;
  Liapis, A.; and Compton, K.
\newblock 2017.
\newblock Mixed-initiative creative interfaces.
\newblock In {\em CHI EA'17: Proceedings of the 2016 CHI Conference Extended
  Abstracts on Human Factors in Computing Systems}.
\newblock ACM.

\bibitem[\protect\citeauthoryear{Guzdial and Riedl}{2016}]{guzdial2016game}
Guzdial, M., and Riedl, M.
\newblock 2016.
\newblock Game level generation from gameplay videos.
\newblock In {\em Twelfth Artificial Intelligence and Interactive Digital
  Entertainment Conference}.

\bibitem[\protect\citeauthoryear{Guzdial \bgroup et al\mbox.\egroup
  }{2017}]{guzdial2017general}
Guzdial, M.; Chen, J.; Chen, S.-Y.; and Riedl, M.~O.
\newblock 2017.
\newblock A general level design editor for co-creative level design.
\newblock {\em Fourth Experimental AI in Games Workshop}.

\bibitem[\protect\citeauthoryear{Liapis, Yannakakis, and
  Togelius}{2013}]{liapis2013sentient}
Liapis, A.; Yannakakis, G.~N.; and Togelius, J.
\newblock 2013.
\newblock Sentient sketchbook: Computer-aided game level authoring.
\newblock In {\em Proceedings of ACM Conference on Foundations of Digital
  Games},  213--220.
\newblock FDG.

\bibitem[\protect\citeauthoryear{Mnih \bgroup et al\mbox.\egroup
  }{2013}]{mnih2013playing}
Mnih, V.; Kavukcuoglu, K.; Silver, D.; Graves, A.; Antonoglou, I.; Wierstra,
  D.; and Riedmiller, M.
\newblock 2013.
\newblock Playing atari with deep reinforcement learning.
\newblock {\em arXiv preprint arXiv:1312.5602}.

\bibitem[\protect\citeauthoryear{Rohanimanesh and
  Mahadevan}{2003}]{rohanimanesh2003learning}
Rohanimanesh, K., and Mahadevan, S.
\newblock 2003.
\newblock Learning to take concurrent actions.
\newblock In {\em Advances in neural information processing systems},
  1651--1658.

\bibitem[\protect\citeauthoryear{Shaker, Shaker, and
  Togelius}{2013}]{shaker2013ropossum}
Shaker, N.; Shaker, M.; and Togelius, J.
\newblock 2013.
\newblock Ropossum: An authoring tool for designing, optimizing and solving cut
  the rope levels.
\newblock In {\em Proceedings of the Ninth AAAI Conference on Artificial
  Intelligence and Interactive Digital Entertainment}.

\bibitem[\protect\citeauthoryear{Smith, Whitehead, and
  Mateas}{2010}]{smith2010tanagra}
Smith, G.; Whitehead, J.; and Mateas, M.
\newblock 2010.
\newblock Tanagra: A mixed-initiative level design tool.
\newblock In {\em Proceedings of the Fifth International Conference on the
  Foundations of Digital Games},  209--216.
\newblock ACM.

\bibitem[\protect\citeauthoryear{Snodgrass and
  Onta{\~n}{\'o}n}{2014}]{snodgrass2014experiments}
Snodgrass, S., and Onta{\~n}{\'o}n, S.
\newblock 2014.
\newblock Experiments in map generation using markov chains.
\newblock In {\em FDG}.

\bibitem[\protect\citeauthoryear{Snodgrass and
  Ontan{\'o}n}{2017}]{snodgrass2017learning}
Snodgrass, S., and Ontan{\'o}n, S.
\newblock 2017.
\newblock Learning to generate video game maps using markov models.
\newblock {\em IEEE Transactions on Computational Intelligence and AI in Games}
  9(4):410--422.

\bibitem[\protect\citeauthoryear{Summerville and
  Mateas}{2016}]{summerville2016super}
Summerville, A., and Mateas, M.
\newblock 2016.
\newblock Super mario as a string: Platformer level generation via lstms.
\newblock In {\em The 1st International Conference of DiGRA and FDG}.

\bibitem[\protect\citeauthoryear{Summerville \bgroup et al\mbox.\egroup
  }{2017}]{summerville2017procedural}
Summerville, A.; Snodgrass, S.; Guzdial, M.; Holmg{\aa}rd, C.; Hoover, A.~K.;
  Isaksen, A.; Nealen, A.; and Togelius, J.
\newblock 2017.
\newblock Procedural content generation via machine learning (pcgml).
\newblock {\em arXiv preprint arXiv:1702.00539}.

\bibitem[\protect\citeauthoryear{Summerville}{2018}]{summerville2018learning}
Summerville, A.
\newblock 2018.
\newblock {\em Learning from Games for Generative Purposes}.
\newblock Ph.D. Dissertation, UC Santa Cruz.

\bibitem[\protect\citeauthoryear{Yannakakis, Liapis, and
  Alexopoulos}{2014}]{yannakakis2014mixed}
Yannakakis, G.~N.; Liapis, A.; and Alexopoulos, C.
\newblock 2014.
\newblock Mixed-initiative co-creativity.
\newblock In {\em Proceedings of the 9th Con- ference on the Foundations of
  Digital Games}.
\newblock FDG.

\bibitem[\protect\citeauthoryear{Zhu \bgroup et al\mbox.\egroup
  }{2018}]{zhuexplainable}
Zhu, J.; Liapis, A.; Risi, S.; Bidarra, R.; and Youngblood, G.~M.
\newblock 2018.
\newblock Explainable ai for designers: A human-centered perspective on
  mixed-initiative co-creation.
\newblock {\em Computational Intelligence in Games}.

\end{thebibliography}

\end{document}